\PassOptionsToPackage{table}{xcolor}
\documentclass[11pt,a4paper,logo,nonumbering]{style}
\usepackage{array,booktabs,tabularx,multirow,graphicx}
\usepackage{tikz,pgfplots}
\pgfplotsset{compat=1.18}
\usepgfplotslibrary{groupplots}
\usetikzlibrary{arrows.meta,positioning,fit,calc,backgrounds,shapes.geometric}
\usepackage{tcolorbox}
\tcbuselibrary{breakable}
\usepackage{adjustbox}
\usepackage{placeins}
\usepackage{fontawesome5}
\definecolor{brandorange}{HTML}{ED6000}
\definecolor{absframe}{HTML}{ED6000}
\definecolor{c1}{HTML}{0072B2}
\definecolor{c2}{HTML}{56B4E9}
\definecolor{c3}{HTML}{009E73}
\definecolor{c4}{HTML}{E69F00}
\definecolor{c5}{HTML}{D55E00}
\definecolor{c6}{HTML}{CC79A7}
\definecolor{r1}{HTML}{6BAED6}
\definecolor{r2}{HTML}{2171B5}
\definecolor{r3}{HTML}{08306B}
\newcommand{\talign}{API-TAIL}
\newcommand{\pool}{\mathcal{C}}
\graphicspath{{assets/}}
\hypersetup{pdftitle={API-TAIL: Detecting Model Substitution in LLM APIs through Temperature-Aligned Likelihood Tails},pdfauthor={TokenRhythm Technologies}}
\makeatletter
\renewcommand{\abscontent}{%
  \begin{tcolorbox}[breakable,colframe=absframe,colback=absframe!5,boxrule=1pt,arc=8pt,
    left=10pt,right=10pt,top=10pt,bottom=10pt]
  \centerline{\fontsize{15pt}{14pt}\selectfont\textbf{Abstract}}\vspace{2ex}
  {\absfont\theabstract\par\vspace{1.2ex}\noindent\centerline{\faGithub\hspace{0.4em}\textbf{Project Page:}~\href{https://github.com/TokenRhythm/API-TAIL}{https://github.com/TokenRhythm/API-TAIL}}}
  \end{tcolorbox}%
}
\renewcommand{\maketitle}{%
  \begingroup\setlength{\parindent}{0pt}
  \vspace*{3pt}
  {\centering\titlefont\@title\par}
  \vskip12pt
  {\centering\@author\par}
  \vskip14pt
  \endgroup
  \thispagestyle{firststyle}\abscontent
}
\makeatother
\fancypagestyle{firststyle}{%
  \fancyhf{}
  \fancyhead[L]{\smash{\raisebox{0.15em}{\includegraphics[height=1.15em]{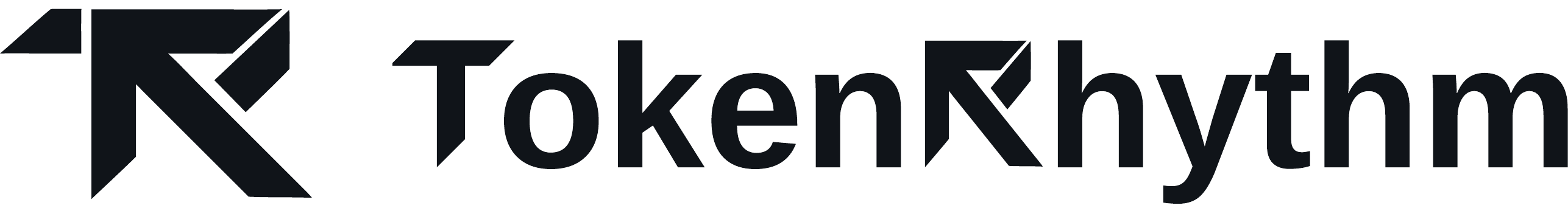}}}}

}
\title{API-TAIL: Detecting Model Substitution in LLM APIs\\through Temperature-Aligned Likelihood Tails}
\renewcommand\Affilfont{\normalfont\fontsize{10}{12}\selectfont\centering}
\author{TokenRhythm Technologies}
\date{}
\begin{abstract}
Commercial language model APIs return only text, allowing a provider to cut costs by quietly routing a fraction of user requests to a cheaper substitute model while billing for the advertised one.
Detecting such partial substitution is challenging because honest providers also alter the output distribution through undocumented decoding and serving choices---such as temperature scaling, sampling truncation, repetition and presence penalties, and default system prompts---forcing an audit to test against a composite null of legitimate configurations rather than a single reference.
Our key insight is that these legitimate variations primarily shift response typicality across the entire batch, whereas partial substitution injects individual responses that remain anomalous under the claimed model.
Building on this, we introduce \talign{}, a typicality audit that first absorbs batch-level shifts by aligning a single effective temperature to center standardized surprisal scores across the audited batch, and then flags the endpoint if the number of remaining tail responses exceeds a threshold calibrated across legitimate configurations.
Evaluated on real-world API responses and constructed substitution mixtures, \talign{} controls false alarms across diverse honest providers and detects low-rate partial substitution with a small query budget. Under a common composite calibration, it outperforms the adapted rank- and MMD-based audit baselines.

\end{abstract}
\begin{document}
\maketitle
\section{Introduction}
\label{sec:intro}

Commercial language model APIs allow users to select and
pay for a specific model by name while returning only generated text. A provider can therefore cut cost by answering a fraction $\rho$
of the requests with a cheaper substitute, such as an older release, a smaller or mixture-of-experts sibling, or a model
of another family, while billing for the advertised model. Such \emph{partial substitution} is invisible in any single
response, and auditing APIs for it has become an active
problem~\citep{gao2025met,cai2025auditing,zhu2025rut,chauvin2026change,zhang2026iris,zhang2026realmoney,lin2026gatescope}.

The central obstacle is that honest providers do not serve a model in one canonical way. They choose decoding
temperatures and truncation thresholds, apply repetition or presence penalties, and add
default system prompts, usually without documenting these choices. Each choice shifts the distribution of the returned
text, so an honest endpoint is not a sample from one reference distribution but from one of many legitimate
deployments: the audit has to test a \emph{composite} null hypothesis. Calibration against a single reference distribution can increase false alarms under legitimate deployment variation,
while recalibration over a broader null can reduce power. On the controlled short-response panels,
RUT~\citep{zhu2025rut} and MET~\citep{gao2025met} baselines calibrated on the same legitimate configurations as
\talign{} have zero power against 10\% substitution with 200 API requests.
These comparisons use method-specific sampling: RUT needs locally generated reference responses for each prompt,
whereas matched-budget MET also needs repeated API responses per prompt (Appendix~\ref{app:baselines}).

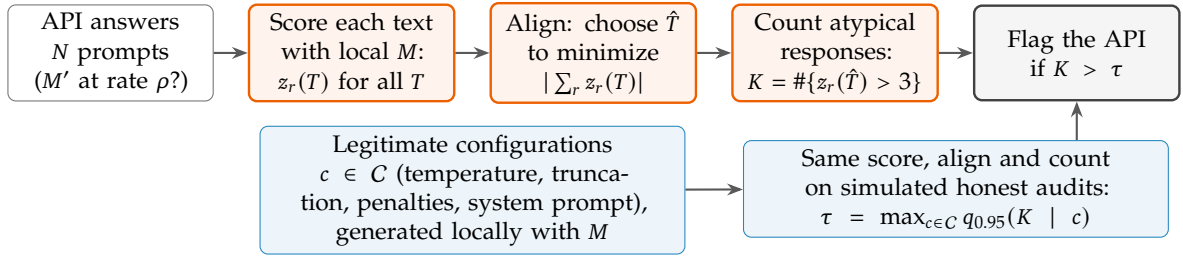
\begin{figure}[t]
\centering
\begin{tikzpicture}[
  font=\footnotesize,
  box/.style={draw=black!55, rounded corners=3pt, align=center, minimum height=1.25cm, text width=2.5cm, fill=white,
              inner sep=3pt},
  hl/.style={box, draw=brandorange, line width=0.9pt, fill=brandorange!7},
  cal/.style={box, draw=c1!85, fill=c1!7, minimum height=1.0cm, text width=5.4cm},
  dec/.style={box, draw=black!75, line width=0.9pt, fill=black!4},
  arr/.style={-{Stealth[length=2.1mm]}, thick, black!65},
  node distance=0.45cm]
\node[box] (api) {API answers\\$N$ prompts\\($M'$ at rate $\rho$?)};
\node[hl, right=of api] (score) {Score each text\\with local $M$:\\$z_r(T)$ for all $T$};
\node[hl, right=of score] (align) {Align: choose $\hat T$\\to minimize\\$|\sum_r z_r(T)|$};
\node[hl, right=of align] (count) {Count atypical\\responses:\\$K=\#\{z_r(\hat T)>3\}$};
\node[dec, right=of count] (flag) {Flag the API\\if $K>\tau$};
\node[cal] (pool) at ($(score.south)!0.5!(align.south)+(0,-1.15)$) {Legitimate configurations $c\in\pool$ (temperature,
truncation, penalties, system prompt), generated locally with $M$};
\node[cal] (thr) at ($(count.south)!0.5!(flag.south)+(0,-1.15)$) {Same score, align and count on simulated honest
audits: $\tau=\max_{c\in\pool} q_{0.95}(K\mid c)$};
\draw[arr] (api) -- (score);
\draw[arr] (score) -- (align);
\draw[arr] (align) -- (count);
\draw[arr] (count) -- (flag);
\draw[arr] (pool) -- (thr);
\draw[arr] (thr.north -| flag.south) -- (flag.south);
\end{tikzpicture}
\caption{\talign{} audits an API from returned text. Each response is scored under the claimed model $M$;
one effective temperature is aligned to the audited batch to absorb batch-level shifts from honest decoding and serving choices; the number of
responses that remain atypical is compared with a threshold calibrated by running the same procedure on locally
generated responses of $M$ across legitimate deployment configurations.}
\label{fig:overview}
\end{figure}

Our approach starts from how the two kinds of change act on the text when every returned response is scored under the
claimed model. Legitimate decoding and serving choices---whether temperature scaling, truncation, repetition and
presence penalties, or default system prompts---shift how surprising the honest responses are for the claimed model, and
they shift the whole batch together. A substitute model instead writes individual responses that stay atypical for the
claimed model across temperatures. This suggests absorbing the batch-level shift first and testing what remains against
a threshold that accounts for residual configuration differences.

\talign{} implements this idea in three steps (Figure~\ref{fig:overview}). It scores each response by its standardized
excess surprisal: how much more surprising the response is under the claimed model at temperature $T$ than a typical
sample, in units of its standard deviation. It \emph{aligns} one effective temperature to the audited batch by choosing
the $T$ at which these scores are centered, thereby absorbing batch-level shifts from undocumented decoding and serving
choices. It then counts the responses whose score stays above $3$ and compares the count with a threshold calibrated on
locally generated responses of the claimed model across a pool of legitimate configurations, including the penalty
configurations that providers use. Scores are computed exactly over the full vocabulary, so an audit is deterministic
and reproducible.

We evaluate \talign{} on real APIs: 26 plain honest endpoints serving Qwen3.8-27B or Gemma-4-31B-it through OpenRouter,
two request protocols, and six substitute models served by seven endpoints, 143{,}955 responses in total. Our
contributions are:
\begin{itemize}[leftmargin=1.4em,itemsep=2pt,topsep=3pt]
\item \textbf{Evaluation setting.} We evaluate text-only API auditing against a composite null spanning legitimate
decoding and serving configurations, with validity measured directly on real honest providers
(Section~\ref{sec:setup}).
\item \textbf{Method.} We propose \talign{}, which aligns an effective temperature by centering standardized typicality
to absorb batch-level deployment shifts, counts the responses that remain atypical, and calibrates the count across a
pool of legitimate decoding and serving configurations (Section~\ref{sec:method}). One audit of 200 requests needs one response per prompt, recorded API charges below $0.16$~USD
and two to eight GPU-minutes of deterministic local scoring.
\item \textbf{Evidence.} \talign{} keeps the false-alarm rate at or below $0.069$ on all 26 plain honest endpoints and
detects 10\% substitution with power of at least $0.69$ in 13 of 16 substitute settings at 200 requests
(Section~\ref{sec:results}). Matched ablations explain the design (Section~\ref{sec:ablation}): alignment versus the
MLE control adds $0.09$ to $0.63$ power by keeping the composite threshold low, full-vocabulary scoring adds
$0.62$ power over a top-20 window on long responses, and the penalty configurations in the pool are what make the
model developer's own endpoint pass.
\end{itemize}

\section{Problem Setup}
\label{sec:setup}

\paragraph{Access model.}
An API advertises a target model $M$ (for example, Qwen3.8-27B) and bills each request as a call to $M$.
The auditor sends prompts $x_1,\dots,x_N$ and observes only the returned texts $y_1,\dots,y_N$; it does not use
log-probabilities, seeds or logit bias from the endpoint, and it does not assume that the endpoint honours sampling
parameters. With access to the weights of $M$, the auditor can run $M$ locally and compute teacher-forced next-token
distributions for any returned text.

\paragraph{Partial substitution.}
A dishonest endpoint answers each request independently with a substitute model $M'\neq M$ with probability $\rho$
and with $M$ otherwise. Substitutes are typically cheaper: an older release of the same family, a smaller or
mixture-of-experts sibling, or a model of another family. Small $\rho$ is the hard and economically relevant regime:
it saves the provider money while leaving most responses untouched.

\paragraph{Legitimate variation and the composite null.}
An honest endpoint serves $M$, but not necessarily the way the auditor would. Providers choose the decoding
temperature, top-$p$ and top-$k$ cutoffs, presence and repetition penalties, default
system prompts and inference backends~\citep{pape2026silent}, usually without documenting these choices. We therefore define honesty through a set $\pool$ of
\emph{legitimate deployment configurations}: the null hypothesis is that there is a configuration $c\in\pool$ under
which $M$ generates the audited responses~\citep{berger1982,acharyya2025composite}. This null is composite. An audit
that compares the endpoint with a single reference deployment of $M$ conflates legitimate deployment choices with
substitution.

\paragraph{Validity and power.}
An audit maps $N$ responses to a binary verdict. We require it to be valid on real traffic: its false-alarm rate, the
probability of flagging an honest endpoint, must stay at most $0.10$ for every honest provider endpoint in the
evaluation, and this verdict must be stable when the calibration samples are re-drawn (we require it in at least 16 of
20 re-drawn calibrations). Among valid audits we compare \emph{power}, the probability of flagging an endpoint that
substitutes at rate $\rho$, at a fixed budget of $N$ requests.

\section{\talign{}: Temperature-Aligned Typicality Auditing}
\label{sec:method}

\talign{} has three parts (Figure~\ref{fig:overview}): it scores every returned response by how typical it is under
the claimed model at a given temperature, it aligns a single effective temperature to minimize the aggregate
standardized score, and it counts the responses that remain atypical, comparing the count with a threshold calibrated
across the legitimate configurations $\pool$ to cover residual configuration differences.

\subsection{Typicality under a tempered model}
For a prompt $x$ and a returned response $y=(y_1,\dots,y_L)$, let $\ell_t\in\mathbb{R}^{|\mathcal V|}$ be the
next-token logits of $M$ at position $t$ (teacher forcing), and let $q_{t,T}=\mathrm{softmax}(\ell_t/T)$ be the tempered
distribution. With the per-token surprisal $s_t(T)=-\log q_{t,T}(y_t)$, its expectation $H_t(T)$ (the entropy of
$q_{t,T}$) and its variance $\sigma^2_t(T)$ under $q_{t,T}$, we define the excess surprisal, its variance and the
standardized typicality score of the response,
\begin{equation}
S(T)=\sum_{t=1}^{L}\bigl(s_t(T)-H_t(T)\bigr),\qquad V(T)=\sum_{t=1}^{L}\sigma^2_t(T),\qquad z(T)=\frac{S(T)}{\sqrt{V(T)}} .
\end{equation}
When each next token is sampled from $q_{t,T}$, the summands $s_t(T)-H_t(T)$ form a martingale difference
sequence with conditional variances $\sigma_t^2(T)$. Under suitable martingale central-limit conditions,
$z(T)$ admits a standard-normal approximation. We calibrate the audit threshold empirically across legitimate configurations. A response from another model may contain tokens that $M$ finds
surprising in context, producing a large positive $z$.

\subsection{Temperature alignment}
An honest provider may alter the output distribution through temperature scaling, truncation, repetition or presence penalties, or a default system prompt, all of which shift the overall typicality of the batch; evaluating $z$ at a fixed reference $T$ confounds substitution with these deployment choices. \talign{} fits one effective temperature to reduce the aggregate shift,
\begin{equation}
\hat T=\operatorname*{arg\,min}_{T\in\mathcal T}\,\Bigl|\sum_{r=1}^{N} z_r(T)\Bigr|,\qquad
\mathcal T=\{0.30,0.31,\dots,1.80\},
\end{equation}
selecting the grid temperature with the smallest absolute aggregate standardized score.

Alignment differs from fitting the temperature by maximum likelihood. Since
$\frac{\partial}{\partial T}\log q_{t,T}(y_t)=\bigl(s_t(T)-H_t(T)\bigr)/T$, the likelihood equation for $T$ is
$\sum_r S_r(T)=\sum_r \sqrt{V_r(T)}\,z_r(T)=0$, weighting each standardized response score by $\sqrt{V_r(T)}$. The MLE control minimizes $|\sum_r S_r(T)|$ over $\mathcal T$,
whereas \talign{} minimizes $|\sum_r z_r(T)|$, giving each response equal weight.
Alignment therefore targets the same standardized scores used in the tail count, keeping null exceedances small for legitimate
configurations that a pure temperature change does not describe, such as repetition penalties or a default system
prompt. Section~\ref{sec:ablation} shows that this is where the two estimators part ways.

\subsection{Counting atypical responses against a composite null}
After alignment, \talign{} tests for an excess of large standardized scores~\citep{donohojin2004}
by counting upper-tail responses,
\begin{equation}
K=\#\{\,r\le N:\ z_r(\hat T)>3\,\}.
\end{equation}

To calibrate $K$, the auditor generates responses of $M$ locally under every configuration $c\in\pool$ on prompts
disjoint from the audit prompts, simulates honest audits of $N$ requests for each $c$, and records the $95$th percentile
$q_{0.95}(K\mid c)$, an integer. The threshold is the largest of these percentiles,
$\tau=\max_{c\in\pool}q_{0.95}(K\mid c)$, and the endpoint is flagged when $K>\tau$. By construction, the false-alarm
rate is at most $5\%$ under every configuration in $\pool$ (up to simulation error); the experiments test whether this
carries over to real honest endpoints, whose configurations the auditor does not know.

\subsection{Exact and reproducible scoring}
\talign{} scores the full vocabulary. For every position, one forward pass of $M$ yields the logits, from which we
compute float32 exponentials and products and accumulate the moment sums in float64: $\sum_v e^{\ell_{t,v}/T}$, $\sum_v e^{\ell_{t,v}/T}\ell_{t,v}$ and
$\sum_v e^{\ell_{t,v}/T}\ell_{t,v}^2$ for all temperatures on the grid; these give $s_t$, $H_t$ and $\sigma^2_t$
over the full vocabulary, without top-$k$ or nucleus truncation. Scoring runs with batch size one and deterministic GPU
kernels, so repeated audits of the same responses are bitwise identical (Appendix~\ref{app:impl}). Auditing one
endpoint with $N=200$ responses takes about two GPU-minutes for 30-token responses and about eight for responses of up
to 400 tokens on a single L20X GPU.

\section{Experimental Setup}
\label{sec:experiments}

\paragraph{Claimed models and honest endpoints.}
We audit two widely served models offered by many independent providers through OpenRouter: Qwen3.8-27B and
Gemma-4-31B-it. From provider metadata, documentation and a short parameter-compliance probe
(Appendix~\ref{app:endpoints}), we separate \emph{plain honest} endpoints, which advertise BF16, FP16 or FP8 weights or
state no precision, document no system-prompt injection and honour the requested temperature, from \emph{special}
endpoints, which serve FP4-class weights, document injecting their own system prompt, or ignore the temperature
parameter. This yields 26 plain honest endpoints (17 for Qwen3.8-27B and 9 for Gemma-4-31B-it), on which validity is
evaluated, and 10 special endpoints (4 and 6), analysed in Appendix~\ref{app:special}.
One additional endpoint, G14, lacks a complete probe and is reported separately using its available P-user responses.

\paragraph{Request protocols and prompts.}
The main study uses two request protocols. \emph{P30} fixes the sampling parameters (temperature 0.5, top-$p$ 1.0, at
most 30 tokens, reasoning disabled) and is the controlled setting; \emph{P-user} sends requests the way a typical user
does, with the provider's default sampling and responses of up to 400 tokens. Crossing the two models with the two
protocols gives four panels. Main-study audit batches are drawn from a held-out pool of 1{,}000 test prompts
(WildChat conversations~\citep{zhao2024wildchat}, code and mathematics, in English and Chinese), disjoint from calibration prompts.

\paragraph{Substitutes and simulated audits.}
We collected real API responses from six substitute models served by seven endpoints: for Qwen3.8-27B, the previous
same-family release Qwen3.6-27B (from a third-party provider, ``provider A'', and from the developer's own platform),
the mixture-of-experts sibling Qwen3.6-35B-A3B, the smaller sibling Qwen3.8-Flash, and DeepSeek-V4.1-Flash; for
Gemma-4-31B-it, the mixture-of-experts sibling Gemma-4-26B-A4B and two models of other families (DeepSeek-V4.1-Flash
and Qwen3.8-27B). The main collection contains 143{,}955 responses (every source has at least 99\% of the requested responses).
The campaign ledger records 29.92~USD in API charges. An honest audit uses $N$ API responses, with prompt allocation
set by the method's sampling protocol; a substitution audit answers each request with the substitute's API response with probability $\rho$ and otherwise with a
response of the claimed model generated locally under the protocol's default configuration. We vary
$\rho\in\{0.05,0.10,0.25\}$ and $N\in\{100,200,400\}$, and simulate 4{,}000 audits per honest endpoint and 1{,}000
audits per substitution setting. Honest-endpoint audits use real API responses and satisfy the false-alarm
criterion in Section~\ref{sec:results}.

\paragraph{Legitimate configurations.}
The main-study pool $\pool$ contains 16--19 configurations per panel, generated locally with $M$: default sampling
(two seeds), temperatures 0.7 and 1.0 with top-$p$ and top-$k$ variants, the model's generation-config defaults, a
default system prompt, presence penalties 1.5 and 2.0, repetition penalties
1.05 and 1.10, one combined penalty setting, and precision variants (Appendix~\ref{app:pool}). Penalties belong in $\pool$ because real
providers apply them; an earlier profiling study of our own found that the model developer's endpoint applies a
presence penalty of about 1.5. Each configuration contributes 2{,}000 simulated honest audits for calibration.

\paragraph{Baselines.}
Matched-budget comparisons use common source response banks, the same legitimate pool and the same
maximum-per-configuration quantile calibration, with method-specific prompt allocation and reference sampling.
We compare adapted implementations of MET~\citep{gao2025met}, a maximum-mean-discrepancy test with a Hamming kernel,
and RUT~\citep{zhu2025rut}, a rank-uniformity test. Matched-budget MET uses two API and two local reference responses
per prompt; RUT uses one API response and 100 local reference responses per prompt. These comparisons run on the
P30 panels; Appendix~\ref{app:baselines} specifies the adaptations.
A separate Qwen check uses native-style MET sampling: $T=1$, 50 tokens, 10 API and 10 local responses per prompt,
and an 11-configuration null pool. Calibration and evaluation independently draw trials from the same 99 eligible prompts.
At $N=200$, we test Qwen3.6-27B (provider A) and DeepSeek-V4.1-Flash substitutions.
We include NLL, log-rank and entropy count controls inspired by text-detection
statistics~\citep{gehrmann2019gltr,mitchell2023detectgpt,su2023detectllm,hans2024binoculars};
each uses the 99th pool percentile as its per-response cutoff.
Section~\ref{sec:ablation} compares ablations of our statistic: fixed-temperature typicality at $T=1$, related by
a sign reversal to same-model Fast-DetectGPT~\citep{bao2024fastdetectgpt}, the MLE temperature control,
and scoring restricted to a top-20 window or the $0.95$ nucleus~\citep{holtzman2020curious}.

\paragraph{Statistical evaluation.}
The pools, thresholds and comparisons were fixed before any collected API response was scored. Confidence intervals for
differences in power use a paired bootstrap over test prompts ($B=2{,}000$) that re-draws the calibration audits and
recalibrates the threshold in every replicate, and the paired comparisons within each panel are corrected with Holm's
method~\citep{holm1979}.

\section{Substitution Detection under Deployment Variation}
\label{sec:results}

\begin{table}[t]
\centering
\small
\setlength{\tabcolsep}{4.2pt}
\begin{tabular}{l cc cc cc cc}
\toprule
 & \multicolumn{2}{c}{Qwen P30} & \multicolumn{2}{c}{Qwen P-user} & \multicolumn{2}{c}{Gemma P30} & \multicolumn{2}{c}{Gemma P-user}\\
\cmidrule(lr){2-3}\cmidrule(lr){4-5}\cmidrule(lr){6-7}\cmidrule(lr){8-9}
Audit & Power\,$\uparrow$ & FAR\,$\downarrow$ & Power\,$\uparrow$ & FAR\,$\downarrow$ & Power\,$\uparrow$ & FAR\,$\downarrow$ & Power\,$\uparrow$ & FAR\,$\downarrow$\\
\midrule
\textbf{\talign{} (ours)} & \textbf{0.33} & 0.035 & \textbf{0.69} & 0.002 & \textbf{0.89} & 0.003 & \textbf{1.00} & 0.069 \\
\midrule
NLL count & 0.00 & 0.000 & 0.00 & 0.000 & 0.08 & 0.000 & 0.13 & 0.000 \\
Log-rank count & 0.00 & 0.000 & 0.00 & 0.000 & 0.03 & 0.000 & 0.10 & 0.000 \\
Entropy count & 0.01 & 0.022 & 0.00 & 0.065 & 0.02 & 0.029 & 0.00 & 0.001 \\
RUT~\citep{zhu2025rut} & 0.00 & 0.003 & \multicolumn{2}{c}{n/a} & 0.00 & 0.000 & \multicolumn{2}{c}{n/a} \\
MET~\citep{gao2025met} & 0.00 & 0.005 & \multicolumn{2}{c}{n/a} & 0.00 & 0.000 & \multicolumn{2}{c}{n/a} \\

\bottomrule
\end{tabular}
\caption{Audits with API budget $N=200$, common source banks, pool $\pool$ and calibration rule,
and method-specific sampling.
\emph{FAR}~($\downarrow$): largest false-alarm rate over the plain endpoints of the panel (17 for Qwen, 9 for Gemma); all values are
below the $0.10$ validity limit and the validity verdicts hold in 20 of 20 re-drawn calibrations. \emph{Power}~($\uparrow$): 10\%
substitution by the panel's main substitute (Qwen3.6-27B served by provider A for Qwen; Gemma-4-26B-A4B for Gemma). RUT and matched-budget MET are evaluated on the P30 panels: RUT uses one API response and 100 local
references per prompt; MET uses two responses per side.}
\label{tab:main}
\end{table}

\paragraph{Validity on real providers.}
Table~\ref{tab:main} first answers whether an audit can be trusted on real traffic. Calibrated on the pool $\pool$, every
audit keeps the false-alarm rate on the evaluated plain endpoints below $0.10$. For \talign{}, the largest false-alarm
rate over the 17 Qwen and 9 Gemma plain endpoints is $0.035$, $0.002$, $0.003$ and $0.069$ on the four panels, and the
validity verdict is the same in all 20 re-drawn calibrations. The endpoints are served by independent providers whose
decoding settings the auditor does not know; the pool covers them through its range of temperatures, cutoffs, penalties
and default prompts.

\paragraph{Detection at the main setting.}
Validity is easy to obtain by flagging nothing, so the comparison that matters is power at equal validity.
\talign{} detects 10\% substitution of the main substitute with power $0.33$ (Qwen P30), $0.69$ (Qwen P-user), $0.89$
(Gemma P30) and $1.00$ (Gemma P-user) at $N=200$. With the common composite calibration,
RUT and matched-budget MET have power $0.00$ in the P30 main settings.
The separate Qwen native-sampling check at $N=200$ also yields power $0.00$ for both tested substitutes
at rates up to $0.25$ (Section~\ref{sec:experiments}).
The likelihood-count tests (NLL, log-rank and entropy counts) reach at most $0.13$ across all four panels. Section~\ref{sec:ablation} further compares \talign{} against ablations of its own alignment and scoring components.

\begin{table}[t]
\centering
\small
\setlength{\tabcolsep}{4.0pt}
\begin{tabular}{ll ccc c}
\toprule
 & & \multicolumn{3}{c}{Power\,$\uparrow$ ($N=200$)} & Power\,$\uparrow$ ($N=400$)\\
\cmidrule(lr){3-5}\cmidrule(lr){6-6}
Panel & Substitute & $\rho=0.05$ & $\rho=0.10$ & $\rho=0.25$ & $\rho=0.10$\\
\midrule
\multirow{5}{*}{Qwen P30} & Qwen3.6-27B (provider A) & 0.09 & 0.33 & 0.73 & 0.62 \\
 & Qwen3.6-27B (developer platform) & 0.11 & 0.42 & 0.78 & 0.71 \\
 & Qwen3.6-35B-A3B & 0.23 & 0.73 & 0.98 & 0.96 \\
 & Qwen3.8-Flash & 0.29 & 0.83 & 1.00 & 0.98 \\
 & DeepSeek-V4.1-Flash & 0.76 & 1.00 & 1.00 & 1.00 \\
\midrule
\multirow{5}{*}{Qwen P-user} & Qwen3.6-27B (provider A) & 0.19 & 0.69 & 0.98 & 0.93 \\
 & Qwen3.6-27B (developer platform) & 0.00 & 0.00 & 0.07 & 0.00 \\
 & Qwen3.6-35B-A3B & 0.74 & 1.00 & 1.00 & 1.00 \\
 & Qwen3.8-Flash & 0.69 & 0.99 & 1.00 & 1.00 \\
 & DeepSeek-V4.1-Flash & 0.94 & 1.00 & 1.00 & 1.00 \\
\midrule
\multirow{3}{*}{Gemma P30} & Gemma-4-26B-A4B & 0.29 & 0.89 & 1.00 & 1.00 \\
 & DeepSeek-V4.1-Flash & 0.56 & 1.00 & 1.00 & 1.00 \\
 & Qwen3.8-27B & 0.54 & 0.99 & 1.00 & 1.00 \\
\midrule
\multirow{3}{*}{Gemma P-user} & Gemma-4-26B-A4B & 0.95 & 1.00 & 1.00 & 1.00 \\
 & DeepSeek-V4.1-Flash & 0.96 & 1.00 & 1.00 & 1.00 \\
 & Qwen3.8-27B & 0.96 & 1.00 & 1.00 & 1.00 \\

\bottomrule
\end{tabular}
\caption{Sensitivity analysis of \talign{}'s detection power~($\uparrow$) across different substitute types (cross-family models, same-family MoE or smaller siblings, and previous same-size releases), substitution rates $\rho$, and query budgets $N$ (calibrated on pool $\pool$; at $N=200$, false-alarm rates are those in Table~\ref{tab:main}). ``Provider A'' and ``developer
platform'' denote two deployments of the same previous-release substitute.}
\label{tab:grid}
\end{table}

\begin{figure}[t]
\centering
\begin{tikzpicture}
\begin{groupplot}[
  group style={group size=4 by 1, horizontal sep=7pt},
  width=0.305\linewidth, height=4.3cm,
  xmode=log, log basis x=2, xtick={100,200,400}, xticklabels={100,200,400},
  xmin=85, xmax=470, ymin=0, ymax=1.04, ytick={0,0.25,0.5,0.75,1},
  grid=major, grid style={black!12}, tick label style={font=\scriptsize},
  title style={font=\small, yshift=-3pt}, xlabel={\scriptsize requests $N$},
  every axis plot/.append style={line width=1pt},
  legend style={font=\scriptsize, draw=none, column sep=6pt}, legend columns=3]
\input{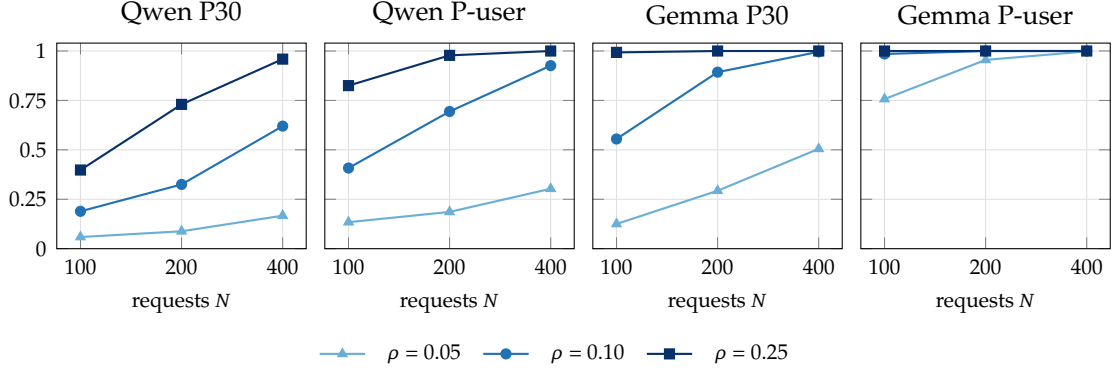}
\end{groupplot}
\end{tikzpicture}\\[3pt]
\ref*{budgetlegend}
\caption{Power of \talign{} as the audit budget grows, for the main substitute of each panel (Qwen3.6-27B
from provider A on the Qwen panels, Gemma-4-26B-A4B on the Gemma panels) at three substitution rates. Every
substitute is listed in Table~\ref{tab:grid} and Appendix~\ref{app:grid}.}
\label{fig:budget}
\end{figure}

\paragraph{Breadth across substitutes.}
Table~\ref{tab:grid} reports every substitute. At $\rho=0.10$ and $N=200$, \talign{} has power of at least $0.69$ in 13
of the 16 substitute--panel settings, and the median over the 16 settings is $0.99$. Models from another family are
detected even at $\rho=0.05$: DeepSeek-V4.1-Flash with power $0.56$ to $0.96$ and Qwen3.8-27B (as a substitute for Gemma)
with $0.54$ and $0.96$. The mixture-of-experts and smaller siblings, which share tokenizer and training lineage with the
claimed model, are detected with power $0.73$ to $1.00$ at $\rho=0.10$. The closest substitute is the previous release
of the claimed model, Qwen3.6-27B: at $\rho=0.10$ it is detected with power $0.33$ (P30) and $0.69$ (P-user) when served
by provider A and with $0.42$ on short responses from the developer's platform, rising to $0.62$, $0.93$ and $0.71$ at
$N=400$. Its developer-platform deployment remains difficult to detect on long default-sampling responses,
with power at most $0.07$ across the evaluated budgets and substitution rates.

\paragraph{Audit budget.}
Power grows smoothly with both the budget and the substitution rate (Figure~\ref{fig:budget}). For the main
substitutes, doubling the budget from $N=200$ to $N=400$ raises power from $0.33$ to $0.62$ on Qwen P30 and from $0.69$ to
$0.93$ on Qwen P-user, and at $\rho=0.25$ already 200 requests give power $0.73$ to $1.00$ on every panel. Over all
substitutes at $N=400$ and $\rho=0.10$, 14 of the 16 settings reach power of at least $0.70$ and 12 reach $0.96$ or
higher (Appendix~\ref{app:grid} gives every budget and rate).

\section{Why the Design Works}
\label{sec:ablation}

\begin{table}[t]
\centering
\footnotesize
\setlength{\tabcolsep}{3.2pt}
\begin{tabular}{l cccc}
\toprule
Variant of \talign{} & Qwen P30 & Qwen P-user & Gemma P30 & Gemma P-user\\
\midrule
MLE temperature & $+0.22$ {\scriptsize [+0.07, +0.27]} & $+0.63$ {\scriptsize [+0.37, +0.67]} & $+0.09$ {\scriptsize [+0.06, +0.13]} & $0.00$ {\scriptsize [0.00, 0.00]} \\
Fixed $T{=}1$, no alignment & $+0.33$ {\scriptsize [+0.22, +0.42]} & $+0.69$ {\scriptsize [+0.47, +0.78]} & $-0.04$ {\scriptsize [-0.07, +0.05]} & $+1.00$ {\scriptsize [+1.00, +1.00]} \\
Top-20 window & $-0.01$ {\scriptsize [-0.04, +0.02]} & $+0.62$ {\scriptsize [+0.41, +0.68]} & $0.00$ {\scriptsize [0.00, 0.00]} & $0.00$ {\scriptsize [0.00, 0.00]} \\
Nucleus ($p{=}0.95$) & $+0.30$ {\scriptsize [+0.20, +0.40]} & $+0.69$ {\scriptsize [+0.47, +0.78]} & $+0.83$ {\scriptsize [+0.78, +0.88]} & $0.00$ {\scriptsize [0.00, +0.01]} \\

\midrule
Threshold $\tau$ (\talign{} / MLE) & 7 / 9 & 5 / 10 & 8 / 9 & 5 / 5 \\

Max.\ FAR without penalties & 0.269$^{\dagger}$ & 0.074 & 0.003 & 0.069 \\

\bottomrule
\end{tabular}
\caption{Design ablations ($N=200$, main substitute, $\rho=0.10$). Top: power of \talign{} minus the
power of the variant, with paired-bootstrap 95\% intervals ($B=2{,}000$); positive values favour \talign{}. Middle:
integer composite-null thresholds $\tau$ of \talign{} and of the MLE variant. Bottom: largest plain-endpoint false-alarm
rate of \talign{} with the five penalty configurations removed from $\pool$ ($^{\dagger}$: above the $0.10$ limit).}
\label{tab:ablation}
\end{table}

Table~\ref{tab:ablation} compares variants of \talign{} on the same responses and pool, with the same calibration
rule applied separately to each variant.

\paragraph{Alignment versus the MLE control.}
The defining choice of \talign{} is its alignment objective. Replacing it by the MLE control, with
everything else fixed, lowers power by $0.22$ on Qwen P30, $0.63$ on Qwen P-user and $0.09$ on Gemma P30; all three
differences are significant after Holm correction, and on Gemma P-user both audits reach power $1.00$. The two
estimators do not disagree about the temperature. From honest traffic to 10\% substitution, the mean fitted temperature
moves from $0.498$ to $0.567$ under alignment and from $0.498$ to $0.561$ under the MLE control on Qwen P30, and from
$0.849$ to $0.869$ versus $0.847$ to $0.867$ on Qwen P-user. They disagree about the null. Because alignment centers the
very standardized scores that are counted, configurations outside the temperature family still produce few
exceedances, and the composite threshold stays low: $7$ versus $9$, $5$ versus $10$ and $8$ versus $9$ atypical
responses out of $200$ on the three unsaturated panels. The configuration that sets the likelihood audit's threshold is
the repetition penalty $1.10$ on Qwen P30, the combined presence and repetition penalty on Qwen P-user, and the default
system prompt on Gemma P30, where the MLE control leaves larger standardized shifts.

\paragraph{Aligning, rather than fixing the temperature.}
Without any alignment, the threshold has to cover every legitimate temperature at once, and power falls to zero on three
panels. Alignment turns these panels into detections with power $0.33$, $0.69$ and $1.00$. On Gemma P30 the two audits are
statistically indistinguishable.

\paragraph{Scoring the full vocabulary.}
Substituted text contains tokens to which the claimed model assigns low probability, and these lie in the tail of the
vocabulary, beyond the top-20 log-probabilities that many serving APIs expose. Computing the scores from a top-20 window
leaves the short-response panels unchanged, but on Qwen P-user, whose responses are up to 400 tokens long, power drops
from $0.69$ to $0.08$. Our nucleus-restricted implementation lowers power by $0.30$ to $0.83$ on the three unsaturated panels.
This difference reflects the joint effect of truncated renormalization and the positions retained for scoring
(Appendix~\ref{app:baselines}).

\paragraph{Covering legitimate penalties.}
Penalties are part of how providers serve models, so they belong in the pool. Without the five penalty
configurations, \talign{} remains valid on three of the four panels, but on Qwen P30 the false-alarm rate of the model
developer's own endpoint rises to $0.269$; including them restores validity, with a maximum false-alarm rate of $0.035$
across Qwen P30 plain endpoints. The endpoint that
documents injecting its own system prompt is flagged in $79\%$ (Qwen P30) and $99\%$ (Qwen P-user) of audits, whereas
FP4-quantized endpoints pass on short responses (Appendix~\ref{app:special}).

\section{Related Work}
\label{sec:related}

\paragraph{Auditing which model an API serves.}
Model equality testing~\citep{gao2025met} compares API and reference samples using maximum mean discrepancy and
also considers a composite null covering FP32 and FP16 reference deployments. The rank-based uniformity
test~\citep{zhu2025rut} uses one API response per natural user prompt and tests its randomized rank among locally
generated reference responses. \citet{cai2025auditing} study model substitution in LLM APIs, including quantized and
randomized substitution, and compare several detection approaches. Empirical studies document deceptive model
claims and routing opacity across commercial API gateways and
providers~\citep{zhang2026realmoney,lin2026gatescope}. Separately, controlled experiments show that inference
backends can change model outputs with fixed weights and decoding parameters~\citep{pape2026silent}.
Related work detects behavioral changes over time~\citep{chen2023chatgpt,richter2025bsa,dima2025changed,chauvin2026change},
fingerprints models or inference stacks using probes or output
idiosyncrasies~\citep{gubri2024trap,finlayson2024logits,pasquini2025llmmap,sun2025idiosyncrasies,hu2025resf,fang2026kbf,bruckner2026onetoken,richardeau2026flips},
audits substitution and routing dilution under a probe budget~\citep{zhang2026iris}, and shows that fingerprints can
be spoofed~\citep{zhang2026spoofing}. With provider cooperation, inference can be verified despite numerical
nondeterminism~\citep{sun2024svip,ong2025toploc,karvonen2025difr}.
\talign{} also operates on natural prompts, using one audited API response per prompt. It combines temperature
alignment with a likelihood-tail count calibrated across legitimate decoding and serving
configurations~\citep{frantar2023gptq,lin2024awq,pape2026silent}, formulated as a composite
null~\citep{berger1982,acharyya2025composite}. Unlike sample-comparison methods such as MET and RUT, it scores
audited responses directly under the claimed model without generating reference responses for each audit prompt.

\paragraph{Likelihood and typicality.}
Likelihood and token-rank statistics support zero-shot text detection~\citep{su2023detectllm};
GLTR~\citep{gehrmann2019gltr} visualizes likelihood, rank and entropy statistics to assist human detection.
DetectGPT~\citep{mitchell2023detectgpt} uses probability curvature;
Fast-DetectGPT~\citep{bao2024fastdetectgpt} forms a standardized log-likelihood criterion using scoring and sampling
models; and Binoculars~\citep{hans2024binoculars} uses the ratio of log-perplexity to log-cross-perplexity.
When both scoring and sampling distributions equal our $q_{t,1}$, the Fast-DetectGPT criterion $d$ satisfies
$z(1)=-d$, apart from numerical variance floors. Our fixed-temperature control counts exceedances in the opposite
score tail and calibrates the batch count across the composite null.
\talign{} instead fits the temperature to the audited batch before counting exceedances~\citep{donohojin2004}.
Comparing surprisal with entropy also underlies locally typical sampling~\citep{meister2023typical};
nucleus sampling~\citep{holtzman2020curious} truncates the vocabulary tail. Our nucleus ablation combines
truncated renormalization with a change in the positions scored (Appendix~\ref{app:baselines}).

\section{Conclusion}
\label{sec:conclusion}

Honest API providers serve the same language model in many legitimate ways, which turns substitution auditing into a
test against a composite null. \talign{} absorbs this variation instead of suffering from it: aligning one effective
temperature absorbs batch-level shifts from honest decoding and serving choices, the responses that remain in the atypical tail carry the evidence of
substitution, and calibrating their count across a pool of legitimate configurations covers residual deployment differences. On real providers of two model families,
the audit meets the false-alarm criterion on all 26 plain honest endpoints and detects 10\% substitution with a few hundred requests,
at recorded API charges of cents and a few GPU-minutes of local scoring.

\section*{Contributors}

\noindent
\mbox{Haocheng Mei},\hspace{0.35em}%
\mbox{Yuchuan Tian},\hspace{0.35em}%
\mbox{Ye Yuan},\hspace{0.35em}%
\mbox{Chao Xu},\hspace{0.35em}%
\mbox{Allen Chen},\hspace{0.35em}%
\mbox{Kai Han},\hspace{0.35em}%
\mbox{Yunhe Wang}\par

\FloatBarrier
\clearpage
\begingroup
\small\setlength{\bibsep}{2.2pt plus 0.5pt}

\endgroup
\clearpage
\appendix
\vspace*{-34pt}
\section{Composite-Null Pool and Endpoints}
\label{app:pool}
\label{app:endpoints}
\label{app:special}
\vspace{-4pt}
\captionsetup{hypcap=false,font=footnotesize,aboveskip=2pt,belowskip=0pt}

\noindent
\begin{minipage}[t]{0.46\linewidth}
\vspace{0pt}
\small
Pool $\pool$ (Table~\ref{tab:pool}) uses local $M$ responses on disjoint prompts
(1{,}000$\times$4 Qwen; 500$\times$2 Gemma), varying temperature ($0.7,1.0$), top-$p$ ($0.8$--$1.0$),
top-$k$ ($\text{none},20,64$), penalties and precision. FP4 or system-prompt endpoints and those failing the
20-prompt probe ($T{=}0$ vs.\ $T{=}1$, first 16 tokens, Fisher $p{<}0.05$) are \emph{special}; other probed
endpoints are \emph{plain}. G14 is unclassified ($^{\ast}$: developer).
\end{minipage}\hfill
\begin{minipage}[t]{0.52\linewidth}
\vspace{-2pt}
\centering
\scriptsize
\renewcommand{\arraystretch}{0.80}
\setlength{\tabcolsep}{2.4pt}
\begin{tabular}{l cccc}
\toprule
Configuration group & Q-P30 & Q-user & G-P30 & G-user\\
\midrule
Protocol defaults (two seeds) & 2 & 2 & 2 & 2\\
Other temperature, top-$p$, top-$k$ & 5 & 8 & 5 & 8\\
Generation-config defaults & 1 & -- & 1 & --\\
Default system prompt & 1 & 1 & 1 & 1\\
FP8 / INT4 weights & 4 & 3 & 2 & 2\\
Penalties (pres., rep., combined) & 5 & 5 & 5 & 5\\
\midrule
Total & 18 & 19 & 16 & 18\\
\bottomrule
\end{tabular}
\captionof{table}{Members of the composite-null pool $\pool$ per panel.}
\label{tab:pool}
\end{minipage}

\vspace{0pt}
{\centering
\scriptsize
\renewcommand{\arraystretch}{0.70}
\setlength{\tabcolsep}{2.2pt}
\begin{tabular}[t]{l l l cc}
\toprule
\multicolumn{5}{c}{Qwen3.8-27B (Q01--Q13)}\\
\midrule
ID & Prec. & Class & P30 & P-user\\
\midrule
Q01 & FP8 & plain & 0.002 & 0.000 \\
Q02$^{\ast}$ & n/s & plain & 0.035 & 0.000 \\
Q03 & FP16 & plain & 0.001 & 0.000 \\
Q04 & FP8 & plain & 0.001 & 0.002 \\
Q05 & n/s & plain & 0.001 & 0.000 \\
Q06 & FP8 & plain & 0.001 & 0.000 \\
Q07 & BF16 & plain & 0.002 & 0.000 \\
Q08 & n/s & plain & 0.002 & 0.000 \\
Q09 & FP8 & plain & 0.004 & 0.000 \\
Q10 & FP8 & plain & 0.002 & 0.000 \\
Q11 & n/s & plain & 0.001 & 0.000 \\
Q12 & FP8 & plain & 0.000 & 0.000 \\
Q13 & n/s & plain & 0.017 & 0.000 \\
\bottomrule
\end{tabular}\hfill
\begin{tabular}[t]{l l l cc}
\toprule
\multicolumn{5}{c}{Qwen (Q14--Q21) \& Gemma (G01--G05)}\\
\midrule
ID & Prec. & Class & P30 & P-user\\
\midrule
Q14 & FP8 & plain & 0.001 & 0.000 \\
Q15 & n/s & plain & 0.000 & 0.000 \\
Q16 & FP8 & plain & 0.001 & 0.000 \\
Q17 & FP8 & plain & 0.002 & 0.000 \\
Q18 & FP4 & FP4 & 0.000 & 0.000 \\
Q19 & FP4 & FP4, no-$T$ & 0.003 & 0.000 \\
Q20 & FP4 & FP4 & 0.000 & 0.000 \\
Q21 & FP8 & sys.\ prompt & 0.788 & 0.993 \\
\midrule
G01 & BF16 & plain & 0.000 & 0.000 \\
G02 & FP8 & plain & 0.002 & 0.000 \\
G03 & FP8 & plain & 0.003 & 0.001 \\
G04 & FP8 & plain & 0.000 & 0.000 \\
G05 & n/s & plain & 0.000 & 0.000 \\
\bottomrule
\end{tabular}\hfill
\begin{tabular}[t]{l l l cc}
\toprule
\multicolumn{5}{c}{Gemma-4-31B-it (G06--G16)}\\
\midrule
ID & Prec. & Class & P30 & P-user\\
\midrule
G06 & BF16 & plain & 0.001 & 0.003 \\
G07 & FP8 & plain & 0.000 & 0.001 \\
G08 & n/s & plain & 0.000 & 0.007 \\
G09 & FP8 & plain & 0.000 & 0.069 \\
G10 & FP4 & FP4, no-$T$ & 0.009 & 0.360 \\
G11 & FP4 & FP4, no-$T$ & 0.012 & 0.344 \\
G12 & FP4 & FP4 & 0.009 & 0.135 \\
G13 & n/s & no-$T$ & 0.004 & 0.134 \\
G14 & n/s & no probe & -- & 0.003 \\
G15 & FP4 & FP4, no-$T$ & 0.013 & 0.023 \\
G16 & FP4 & FP4, sys., no-$T$ & 0.004 & 0.525 \\
\bottomrule
\end{tabular}\par
\captionof{table}{Per-endpoint false-alarm rate of \talign{} ($N{=}200$, pool $\pool$, 4{,}000 audits; ``no-$T$'': $T$ ignored).}
\label{tab:endpoints}
\par}

\vspace{-10pt}
\section{Baselines, Full Power Grid and Reproducibility}
\label{app:baselines}
\label{app:grid}
\label{app:impl}

{\small\setlength{\parskip}{2pt}
MLE minimizes $|\sum_r S_r(T)|$ on $\mathcal T$; fixed-$T$ counts $z_r(1)>3$.
RUT~\citep{zhu2025rut} scores mean log token midranks: FP16 ties within the stored top-$k$ window, reported ranks outside.
Response ranks use $(a+U(b+1))/(m+1)$ instead of $(a+Ub)/m$:
$a,b$ count smaller/equal reference scores, $m=100$, $U\sim\mathrm{Unif}(0,1)$.
The Cram\'er--von Mises $-\log p$ is composite-calibrated.\par
MET~\citep{gao2025met} uses unbiased Hamming MMD: sequences truncated/padded to 30 tokens, two API and two local responses
per prompt, with composite calibration. Native MET: Section~\ref{sec:experiments}.
Nucleus scoring renormalizes the shortest probability-ordered prefix reaching mass $0.95$, capped at 4{,}096 tokens;
insufficient mass retains this cap and records overflow.
Tokens outside contribute to neither $S$ nor $V$, changing both normalization and scored positions.\par
Scoring is bitwise reproducible: Transformers, batch 1, deterministic kernels, FP64 accumulation over 151 temperatures/pass.
Per audit: recorded API charges $0.007$--$0.155$~USD; local scoring 2--8 L20X GPU-minutes.\par}

\vspace{0pt}
{\centering
\scriptsize
\renewcommand{\arraystretch}{0.70}
\setlength{\tabcolsep}{3.4pt}
\begin{tabular}{ll ccc ccc ccc}
\toprule
 & & \multicolumn{3}{c}{$\rho=0.05$} & \multicolumn{3}{c}{$\rho=0.10$} & \multicolumn{3}{c}{$\rho=0.25$}\\
\cmidrule(lr){3-5}\cmidrule(lr){6-8}\cmidrule(lr){9-11}
Panel & Substitute & 100 & 200 & 400 & 100 & 200 & 400 & 100 & 200 & 400\\
\midrule
\multirow{5}{*}{Qwen P30} & Qwen3.6-27B (provider A) & 0.06 & 0.09 & 0.17 & 0.19 & 0.33 & 0.62 & 0.40 & 0.73 & 0.96 \\
 & Qwen3.6-27B (developer platform) & 0.07 & 0.11 & 0.22 & 0.26 & 0.42 & 0.71 & 0.54 & 0.78 & 0.98 \\
 & Qwen3.6-35B-A3B & 0.14 & 0.23 & 0.49 & 0.47 & 0.73 & 0.96 & 0.85 & 0.98 & 1.00 \\
 & Qwen3.8-Flash & 0.20 & 0.29 & 0.57 & 0.53 & 0.83 & 0.98 & 0.93 & 1.00 & 1.00 \\
 & DeepSeek-V4.1-Flash & 0.52 & 0.76 & 0.96 & 0.97 & 1.00 & 1.00 & 1.00 & 1.00 & 1.00 \\
\midrule
\multirow{5}{*}{Qwen P-user} & Qwen3.6-27B (provider A) & 0.13 & 0.19 & 0.30 & 0.41 & 0.69 & 0.93 & 0.82 & 0.98 & 1.00 \\
 & Qwen3.6-27B (developer platform) & 0.00 & 0.00 & 0.00 & 0.01 & 0.00 & 0.00 & 0.04 & 0.07 & 0.07 \\
 & Qwen3.6-35B-A3B & 0.49 & 0.74 & 0.92 & 0.94 & 1.00 & 1.00 & 1.00 & 1.00 & 1.00 \\
 & Qwen3.8-Flash & 0.40 & 0.69 & 0.88 & 0.88 & 0.99 & 1.00 & 1.00 & 1.00 & 1.00 \\
 & DeepSeek-V4.1-Flash & 0.73 & 0.94 & 1.00 & 0.99 & 1.00 & 1.00 & 1.00 & 1.00 & 1.00 \\
\midrule
\multirow{3}{*}{Gemma P30} & Gemma-4-26B-A4B & 0.12 & 0.29 & 0.51 & 0.56 & 0.89 & 1.00 & 0.99 & 1.00 & 1.00 \\
 & DeepSeek-V4.1-Flash & 0.33 & 0.56 & 0.86 & 0.91 & 1.00 & 1.00 & 1.00 & 1.00 & 1.00 \\
 & Qwen3.8-27B & 0.28 & 0.54 & 0.78 & 0.87 & 0.99 & 1.00 & 1.00 & 1.00 & 1.00 \\
\midrule
\multirow{3}{*}{Gemma P-user} & Gemma-4-26B-A4B & 0.76 & 0.95 & 1.00 & 0.98 & 1.00 & 1.00 & 1.00 & 1.00 & 1.00 \\
 & DeepSeek-V4.1-Flash & 0.78 & 0.96 & 1.00 & 0.99 & 1.00 & 1.00 & 1.00 & 1.00 & 1.00 \\
 & Qwen3.8-27B & 0.77 & 0.96 & 1.00 & 0.99 & 1.00 & 1.00 & 1.00 & 1.00 & 1.00 \\

\bottomrule
\end{tabular}\par
\captionof{table}{Power of \talign{} across substitutes, rates $\rho$ and budgets $N\in\{100,200,400\}$ (1{,}000 audits).}
\label{tab:fullgrid}
\par}

\end{document}